\documentclass{article}
\PassOptionsToPackage{square, numbers}{natbib}
\usepackage[final]{nips_2017}
\usepackage[utf8]{inputenc} 
\usepackage[T1]{fontenc}    
\usepackage{hyperref}       
\usepackage{url}            
\usepackage{booktabs}       
\usepackage{amsfonts}       
\usepackage{nicefrac}       
\usepackage{microtype}      
\usepackage{graphicx}
\usepackage{enumitem}
\usepackage{array}
\newcolumntype{P}[1]{>{\centering\arraybackslash}p{#1}}
\usepackage[font=small,labelfont=bf,center]{caption}
\hypersetup{colorlinks=true,urlcolor=blue,linkcolor=black,citecolor=black}

\title{GANime: Generating Anime and Manga Character Drawings from Sketches with Deep Learning}

\author{
  Tai Vu \\
  Department of Computer Science\\
  Stanford University\\
  \texttt{taivu@stanford.edu} \\
  \And
  Robert Yang \\
  Department of Computer Science\\
  Stanford University\\
  \texttt{bobyang9@cs.stanford.edu} \\
}

\begin{document}


\maketitle

\begin{abstract}

The process of generating fully colorized drawings from sketches is a large, usually costly bottleneck in the manga and anime industry. In this study, we examine multiple models for image-to-image translation between anime characters and their sketches, including Neural Style Transfer, C-GAN, and CycleGAN. By assessing them qualitatively and quantitatively, we find that C-GAN is the most effective model that is able to produce high-quality and high-resolution images close to those created by humans.

\end{abstract}

\section{Introduction}

Colorization is an especially prominent bottleneck in manga and anime production. Given busy schedules of monthly releases, many artists opt for black and white images rather than high-quality colorized images due to the lack of personnel to colorize, the lower cost involved, and the faster production rate. Therefore, we seek to design a system that can enable artists to automate their colorization process and speed up production.

In this project, we propose a computer vision program for image-to-image translation between anime characters and their sketches. Specifically, the outcome is a neural network that takes as inputs sketch drawings of anime characters and produces high-resolution color images of these characters.

To attain this outcome, we develop three computer vision models, including Neural Style Transfer \cite{styleTransfer}, Conditional Generative Adversarial Networks (C-GAN) \cite{cGAN} \cite{cGAN2}, and Cycle Generative Adversarial Networks (CycleGAN) \cite{cycleGAN}. These algorithms have demonstrated great success on a number of image translation tasks, so they are appropriate for our project. By employing several qualitative and quantitative evaluation metrics, we find that both the GAN models outperform the baseline. In particular, C-GAN with an added total variation loss performs the best with an average FID of 220.499 and an average SSIM of 0.7559.

\section{Related Work}

Image-to-image translation is an active area of research in computer vision. One popular algorithm is Neural Style Transfer \cite{styleTransfer}, which blends a content image and a style image. It aims to generate an image that is similar to the content image but follows the style of the style image.

Another common solution is Conditional GAN (C-GAN) \cite{cGAN}, a variation of Generative Adversarial Networks \cite{gan} that generates new images based on inputs to the network. In particular, Pix2Pix \cite{cGAN2} is a C-GAN that provides a general-purpose solution to image-to-image translation problems, where we condition on an input image and generate a corresponding output image. These models not only learn the mapping from inputs to outputs, but also learn a loss function to train this mapping. 

Additionally, CycleGAN \cite{cycleGAN} is a state-of-the-art model that learns to translate an image from a source domain to a target domain in the absence of aligned image pairs. It captures special characteristics of the source image collection and translates these characteristics into the other image collection.

There have been some attempts to apply GANs in image colorization problems \cite{cGAN2} \cite{colorize} \cite{colorize2}. However, these papers focused on producing real-life color images from their gray-scale versions. There is little work on handling sketch drawings and fictional cartoon characters (such as \cite{animeColorize}). This is an interesting challenge because sketches contain less rich information than gray-scale inputs. In this paper, we scrutinize the effectiveness of different generative models on line art colorization.

\section{Dataset and Input Pipeline}

We used the Anime Sketch Colorization Pair dataset \cite{data1} from Kaggle. This dataset contains 17769 pairs of sketch-color anime character images, which are separated into 14224 examples for training and 3545 instances for testing. Each of these images is an RGB image of size $512 \times 1024$.

\begin{figure}[!h]
    \centering
    \begin{minipage}[b]{\textwidth}
    \centering
    \includegraphics[width = 0.7\textwidth]{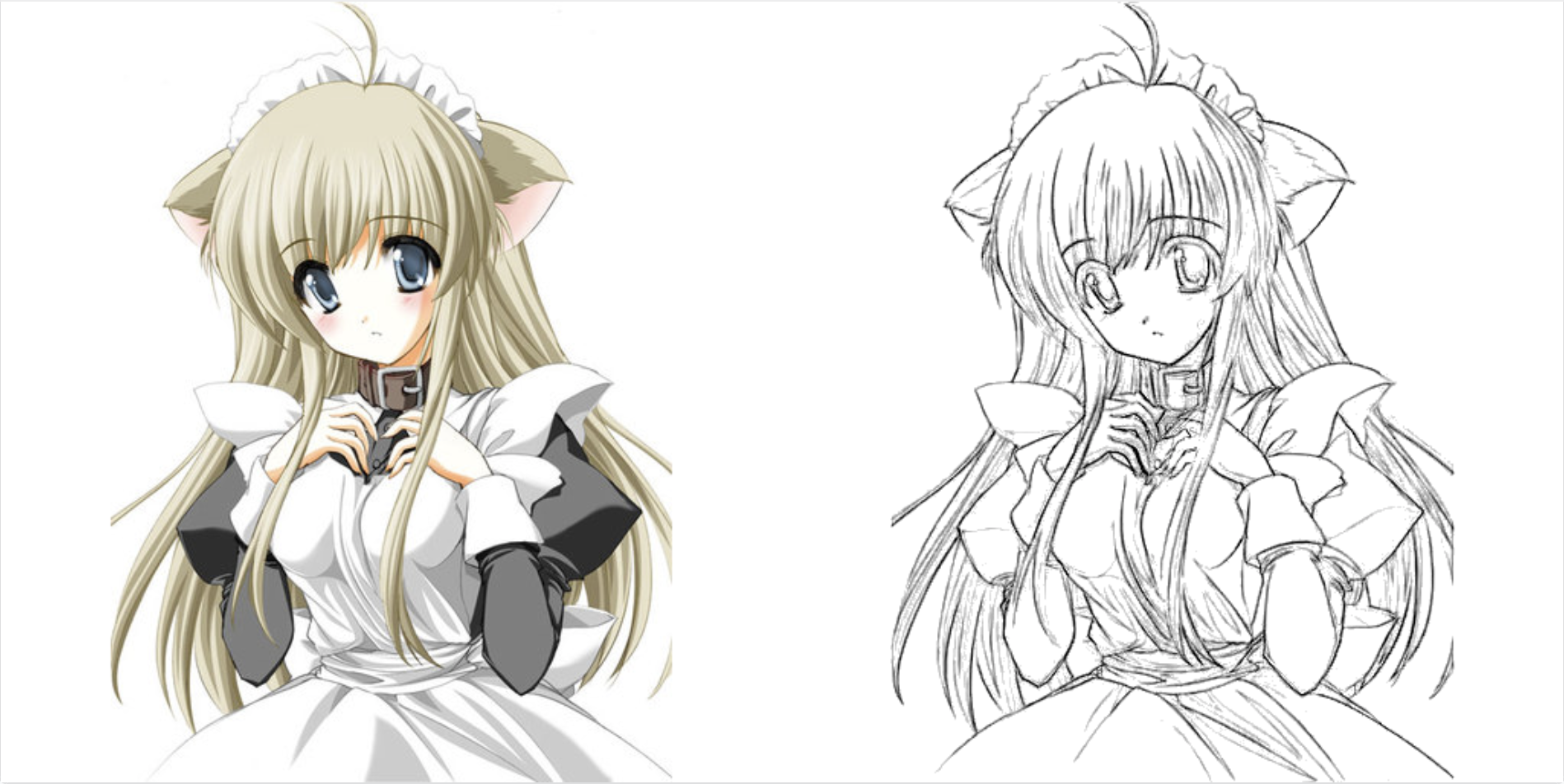}
    \caption{An image in the dataset. \\ Training example (right) and ground truth (left) in the same image.}
    \label{fig:data}
    \end{minipage}
\end{figure}

In the dataset, the training example (a sketch drawing) and the ground truth (a colorized painting) are part of the same image (see Figure \ref{fig:data}), so we ran a script to separate the data into inputs X and outputs Y. We then rescaled the image to $256 \times 256$ resolution, which is a high-enough resolution to not obscure detail, but a low-enough resolution to save memory and expedite the running of the algorithm. In addition, we normalized the training examples to the range $[-1, 1]$, because we believe that normalization can help speed up learning. 

For training GANs, we utilized a batch size of 32 and shuffled the training instances in every epoch. We also implemented some data augmentation techniques, including random cropping and random mirroring.

\section{Implementation}

We leveraged TensorFlow 2.1.0 \cite{tf} for implementing our program. In addition, we harnessed Amazon Web Services (AWS) \cite{aws} and Google Colab with GPU \cite{colab} for training the models.

Link to our project: \url{https://github.com/taivu1998/GANime}

\section{Models}

\subsection{Baseline Model: Neural Style Transfer}

We utilized the Neural Style Transfer \cite{styleTransfer} algorithm as a baseline model. The training example is inputted as the "content" image, and the ground truth is inputted as the "style" image. We observe if the style (colors) from the ground truth can be transferred onto the generated image. 

We designed two different implementations for this task. The first one is Fast Style Transfer \cite{styleTransfer3}, provided by Arbitrary Image Stylization on Tensorflow Hub \cite{tf_hub}. This practice allows for fast real-time stylization with any content-style image pair. The second model is the original style-transfer algorithm with a pretrained VGG19 network. We leveraged its intermediate layers to get the content and style representations and matched our output image to these representations. We trained the model for 1000 epochs.

\subsection{C-GAN}
Next, we built the Pix2Pix model \cite{cGAN2}. A generator $G$ generates a colorized image from an input sketch image, while a discriminator $D$ takes as inputs a sketch image and a color image and determines whether the color image is real or fake. The generator $G$ is a U-Net architecture, which is an encoder-decoder with skip connections from encoder layers to decoder layers. The discriminator $D$ is a PatchGAN architecture, which classifies $N \times N$ patches of an image as real or fake.

The loss function for Pix2Pix is 
$$\mathcal{L}_{cGAN}(G, D) = \mathbb{E}_{x, y}[\log D(x, y)] + \mathbb{E}_{x, z}[\log (1 - D(x, G(x, z)))]$$ where $x$ is an input, $y$ is the corresponding target, and $z$ is a random noise. $G$ tries to minimize this objective against $D$ that tries to maximize it. 

In addition, we added a $L_1$ reconstruction loss 
$$\mathcal{L}_{L_1}(G) = \mathbb{E}_{x, y, z}[\|y - G(x, z)\|_1]$$ 
Hence, the total loss for Pix2Pix is 
$$\mathcal{L} (G, D) = \mathcal{L}_{cGAN}(G, D) + \lambda_{L_1} \mathcal{L}_{L_1}(G)$$

Our objective is to seek $G^* = \arg \min_G \max_D \mathcal{L} (G, D)$.

Furthermore, we invented a new variant of Pix2Pix to enhance its performance. Particularly, we added a total variation loss
$$\mathcal{L}_{tv}(G) = \sum_{i, j} \left(|\hat{y}_{i + 1, j} - \hat{y}_{i, j}| + |\hat{y}_{i, j + 1} - \hat{y}_{i, j}|\right)$$
where $\hat{y} = G(x, z)$. This total variation loss acts as a regularization term that encourages adjacent pixels to have similar values and reduces high frequency artifacts of the generated image.

The modified loss function is
$$\mathcal{L} (G, D) = \mathcal{L}_{cGAN}(G, D) + \lambda_{L_1} \mathcal{L}_{L_1}(G) + \lambda_{tv} \mathcal{L}_{tv}(G)$$

We chose $\lambda_{L_1} = 100, \lambda_{tv} = 0.0001$ and patch size $N = 70$. We trained the model for $150$ epochs using the Adam optimization algorithm with learning rate $\alpha = 0.0002$, $\beta_1 = 0.5, \beta_2 = 0.999$ and $\epsilon = 10^{-7}$.

\begin{figure}[!h]
    \centering
    \begin{minipage}[b]{\textwidth}
    \centering
    \includegraphics[width = \textwidth]{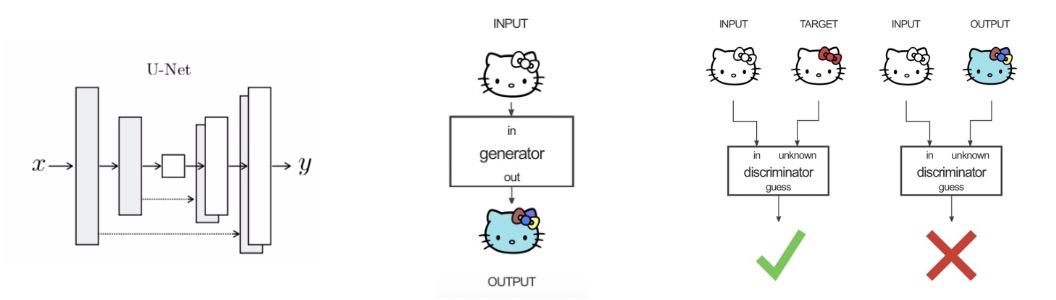}
    \caption{U-Net architecture (left), C-GAN generator (middle), C-GAN discriminator (right) \cite{cGAN2}, \cite{cGAN3}}
    \label{fig:pix2pixarch}
    \end{minipage}
\end{figure}

\subsection{CycleGAN}

Finally, we built a CycleGAN model based on the Pix2Pix architecture design. A generator $G$ generates a colorized image from a sketch image, while a generator $F$ generates a sketch image from a colorized image. Two discriminators $D_X$ and $D_Y$ distinguish between real and fake images. The generators $G$ and $F$ are U-Nets, while the discriminators $D_X$ and $D_Y$ are PatchGANs. 

The loss function for CycleGAN is 
$$\mathcal{L}_{GAN} (G, D_Y, X, Y) = \mathbb{E}_y [\log D_Y(y)] + \mathbb{E}_x [\log(1 - D_Y(G(x)))]$$ where $x$ is an input and $y$ is the corresponding target. $G$ aims to minimize this objective against $D_Y$ that tries to maximize it. Similarly, we have another loss $\displaystyle \mathcal{L}_{GAN} (F, D_X, Y, X)$. 

Additionally, we added a cycle consistency loss 
$$\mathcal{L}_{cyc} (G, F) = \mathbb{E}_x [\|F(G(x)) - x\|]_1 + \mathbb{E}_y [\|G(F(x)) - y\|]_1$$
Hence, the total loss for CycleGAN is 
$$\mathcal{L}(G, F, D_X, D_Y) = \mathcal{L}_{GAN} (G, D_Y, X, Y) + \mathcal{L}_{GAN} (F, D_X, Y, X) + \lambda_{cyc} \mathcal{L}_{cyc} (G, F)$$
Our objective is to seek $\displaystyle G^*, F^* = \arg \min_{G, F} \max_{D_X, D_Y} \mathcal{L}(G, F, D_X, D_Y)$.

We chose $\lambda_{cyc} = 10$ and patch size $N = 70$. We trained the model for $150$ epochs using the Adam optimization algorithm with learning rate $\alpha = 0.0002$, $\beta_1 = 0.5, \beta_2 = 0.999$ and $\epsilon = 10^{-7}$.

\begin{figure}[!h]
    \centering
    \begin{minipage}[b]{\textwidth}
    \centering
    \includegraphics[width = 0.9\textwidth]{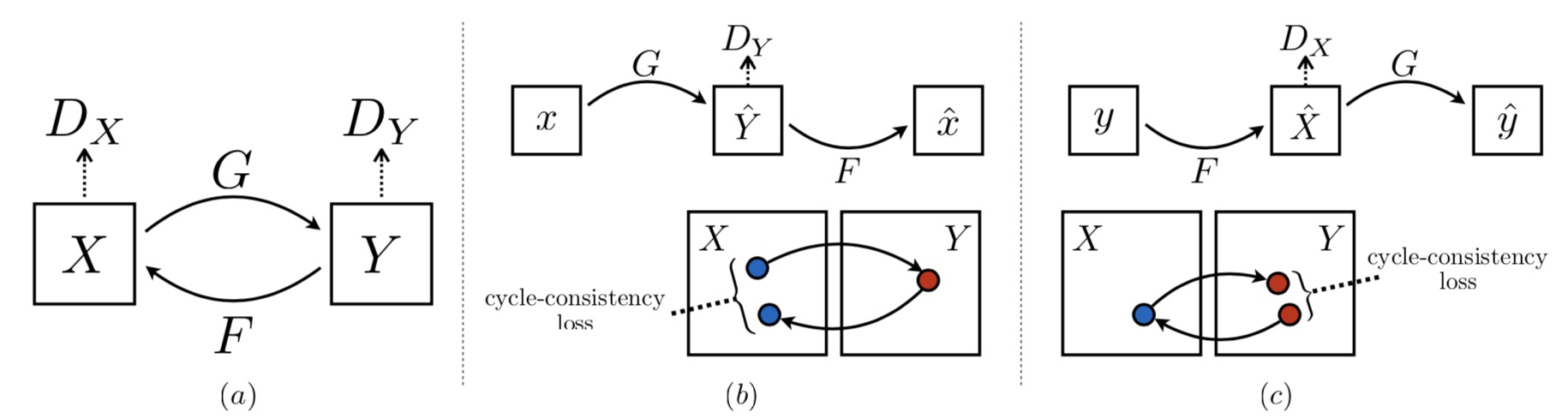}
    \caption{\centering CycleGAN generators and discriminators \cite{cycleGAN}}
    \label{fig:cycleganarch}
    \end{minipage}
\end{figure}

\section{Evaluation}

We evaluated the performance of the three models qualitatively through direct observation and quantitatively using the Structural Similarity (SSIM) Index and the Frechet Inception Distance (FID). The SSIM quantitatively determines the quality of generated outputs by evaluating the difference between the image we want to measure (in this case the generated image) and a good quality image (in this case the ground truth). Meanwhile, the FID \cite{fid} computes the covariance between real and fake distributions using a pre-trained Inception v3 network.

We calculated the average SSIM and the average FID scores for Neural Style Transfer, C-GAN, and CycleGAN, by generating a sample of 100 images from each model and finding the mean and standard deviation of each sample. This informed us on not only the overall quality of the images but also the consistency of the models. We also computed the SSIM and the FID for the two C-GAN versions across various epochs to examine the progress of the C-GAN.

\section{Experiments and Discussion}

\subsection{Qualitative Results}

\subsubsection{Visual Results of Neural Style Transfer}

\begin{figure}[!h]
    \centering
    \begin{minipage}[b]{\textwidth}
    \centering
    \includegraphics[width = \textwidth]{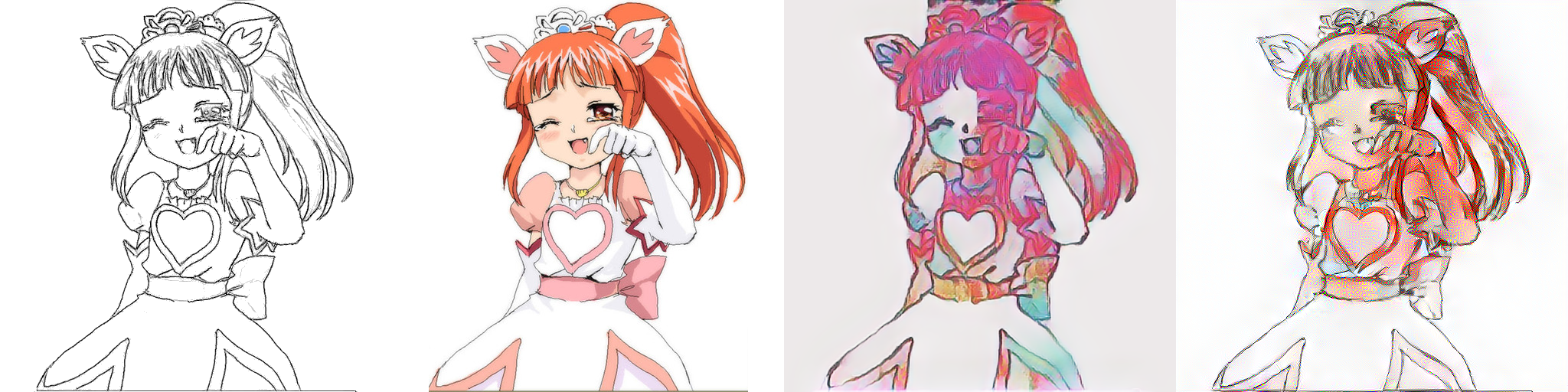}
    \caption{Neural Style Transfer results. Training example (first), ground truth (second), \\ generated image - 1st model (third), generated image - 2nd model (fourth).}
    \label{fig:styleTransfer}
    \end{minipage}
\end{figure}

We can see that overall, the same colors from the style image have been transferred onto the generated image, but the location of those colors are different than expected. More interestingly, the first implementation produces an output image with many distinct colors, while the second one focuses on transferring the main color of the style image to the generated outcome (as observed in Figure \ref{fig:styleTransfer}). We believe that Neural Style Transfer is a good baseline model that our final algorithm should outperform.

\subsubsection{Visual Results of CycleGAN}

\begin{figure}[!h]
    \centering
    \begin{minipage}[b]{\textwidth}
    \centering
    \includegraphics[width = 0.85\textwidth]{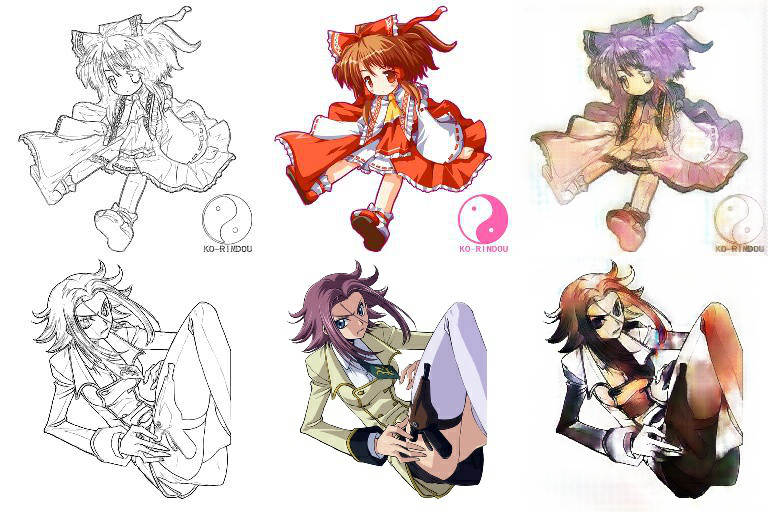}
    \caption{CycleGAN results, epoch 150 \\ Training example (first), ground truth (second), generated image (third).}
    \label{fig:cycleGAN2}
    \end{minipage}
\end{figure}

Overall, we see that the CycleGAN performs slightly better than the baseline from a visual standpoint, but it is far from perfect. It focuses on one or two colors instead of learning to produce different colors, as shown in Figure \ref{fig:cycleGAN2}. In addition, the colorization seems a bit inconsistent and unsmooth, with some portions colorized while others left with no colors.

\subsubsection{Visual Results of C-GAN}
\begin{figure}[!h]
    \centering
    \begin{minipage}[b]{\textwidth}
    \centering
    \includegraphics[width = \textwidth]{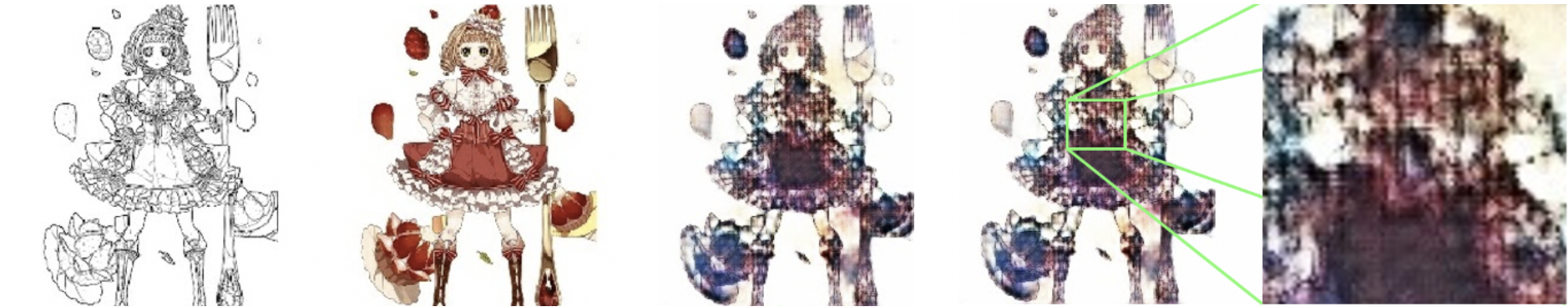}
    \caption{C-GAN results, epoch 2. Training example (first), ground truth (second), \\ generated image (third $\&$ fourth) and close-up of generated image (fifth).}
    \label{fig:cgan_1}
    \end{minipage}
\end{figure}

In the first few epochs, the generated images are often low-quality (see Figure \ref{fig:cgan_1}). Specifically, there is a grainy texture that makes the image look artificial, showed in the close-up. What is more, there can be two colors that are used to color one area which is supposed to be colored by one single color.

\begin{figure}[!h]
    \centering
    \begin{minipage}[b]{\textwidth}
    \centering
    \includegraphics[width = 0.93\textwidth]{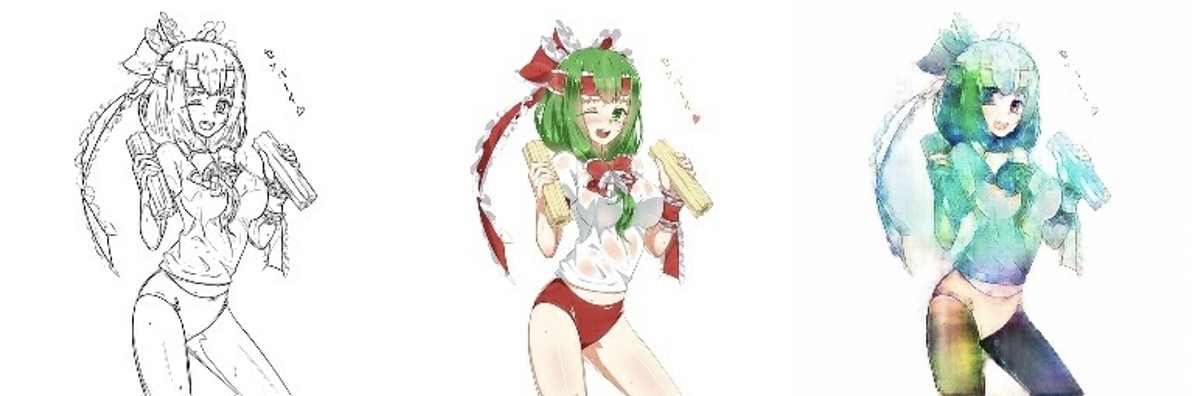}
    \caption{C-GAN results, epoch 18 \\ Training example (first), ground truth (second), generated image (third).}
    \label{fig:cgan_2}
    \end{minipage}
\end{figure}

By epoch 18, the GAN has improved significantly, getting rid of the grainy texture. However, the color problem is still there (as shown in Figure \ref{fig:cgan_2}), where green and purple compete for the same area. 

\begin{figure}[!h]
    \centering
    \begin{minipage}[b]{\textwidth}
    \centering
    \includegraphics[width = 0.85\textwidth]{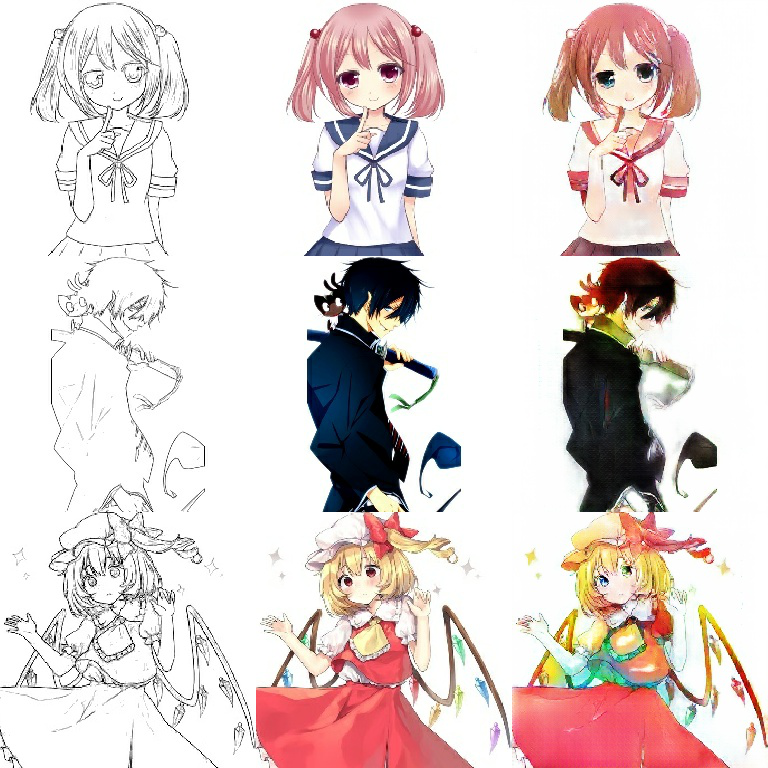}
    \caption{C-GAN results, epoch 150 \\ Training example (first), ground truth (second), generated image (third).}
    \label{fig:cgan_3}
    \end{minipage}
\end{figure}

Afterwards, by epoch 150, the generated images are approaching the qualities of the ground truth images (as observed in Figure \ref{fig:cgan_3}), with the texture problem gone entirely and the coloring problem mostly gone.

\begin{figure}[!h]
    \centering
    \begin{minipage}[b]{\textwidth}
    \centering
    \includegraphics[width = 0.85\textwidth]{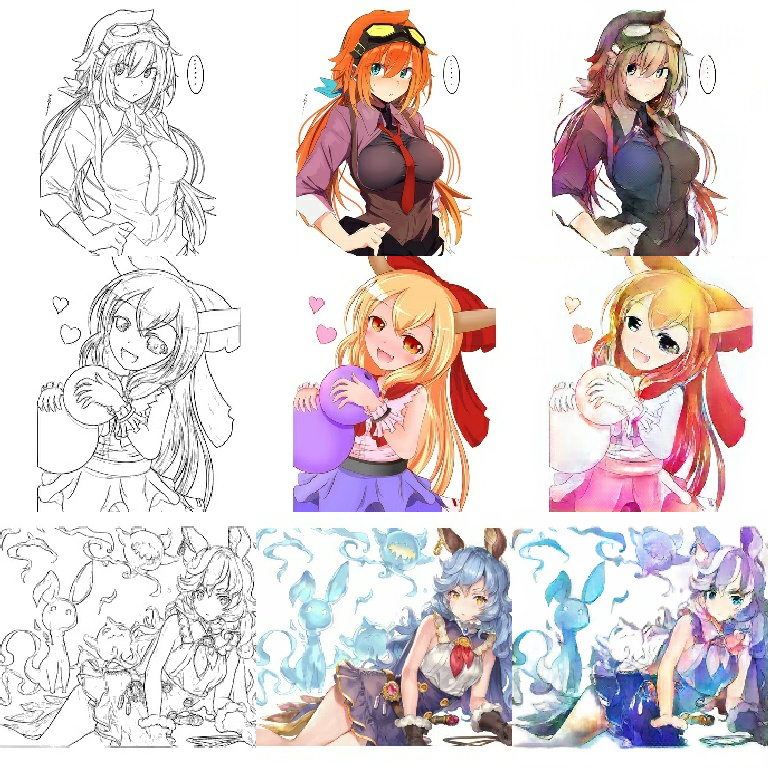}
    \caption{Modified C-GAN results, epoch 150 \\ Training example (first), ground truth (second), generated image (third).}
    \label{fig:cgan_4}
    \end{minipage}
\end{figure}

Qualitatively, the C-GAN (Pix2Pix) excels at coloring the hair of anime characters, often being quite detailed. Furthermore, C-GANs trained on more epochs grow increasingly good at coloring skin and eyes the right color (a rather difficult task for the computer). Lastly, the C-GAN can usually capture the nuances in the character's clothes and accessories by coloring them correctly and beautifully. However, Pix2Pix is still imperfect with sophisticated images with many details, and sometimes still smudges dark colors. Despite that, the C-GAN has very high performance overall. 

Meanwhile, the addition of the total variation loss tends to improve the performance of Pix2Pix. By comparing Figure \ref{fig:cgan_3} and Figure \ref{fig:cgan_4}, we can see that the outputs in the latter one are slightly better looking. In fact, the total variation loss removes many high frequency components and encourages the network to color one region with one single color. Hence, the final outcomes become much smoother.

\subsection{Quantitative Results}

We first compare the performance of Neural Style Transfer, C-GAN, and CycleGAN. In Table \ref{fig:table_fid_ssim}, we see that C-GAN has the best performance according to both the FID and SSIM metrics. Namely, C-GAN has the lowest FID, meaning that the distribution of C-GAN-generated images is closest to the ground truth distribution. It also has the highest average SSIM, meaning that it has the highest overall image quality compared to CycleGAN and Neural Style Transfer. Furthermore, it has the lowest SSIM standard deviation, which signifies superior robustness to input image complexity.

In addition, the modified version of C-GAN has a slightly better performance than the original one. In particular, in comparison with the traditional Pix2Pix, the new one with a total variation loss has a lower FID score, a higher SSIM mean and a lower SSIM standard deviation. This is consistent with our above observation of visual results.

\begin{figure}[ht]
  {
	\begin{minipage}[c][\width]{
	   0.5\textwidth}
	   \centering
	   \includegraphics[width=\textwidth]{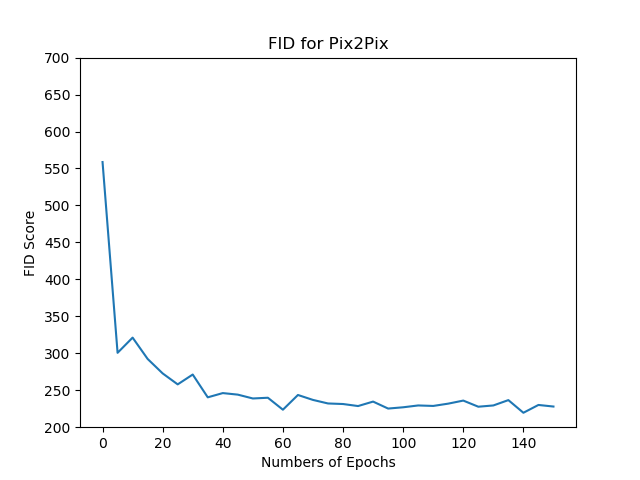}
	\end{minipage}}
  {
	\begin{minipage}[c][\width]{
	   0.5\textwidth}
	   \centering
	   \includegraphics[width=\textwidth]{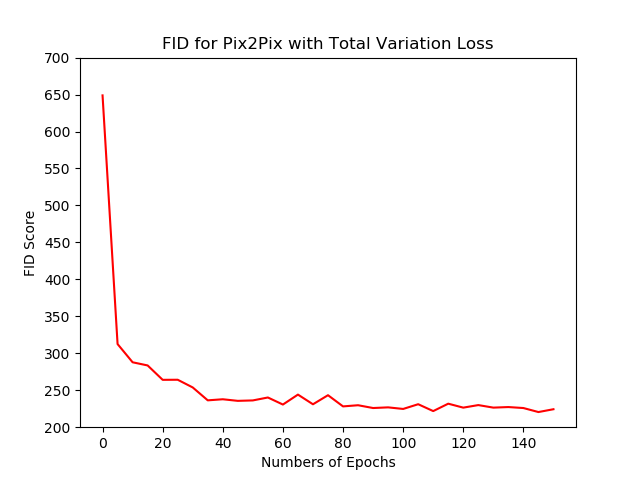}
	\end{minipage}}
\caption{FID scores of C-GAN models trained to epochs 0-150 \\ Pix2Pix (left), modified Pix2Pix (right).}
\label{fig:fid_pix2pix_both}
\end{figure}

\begin{figure}[ht]
  {
	\begin{minipage}[c][\width]{
	   0.5\textwidth}
	   \centering
	   \includegraphics[width=\textwidth]{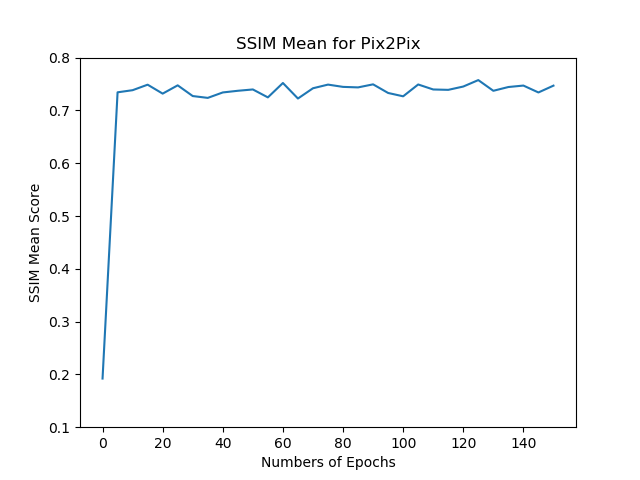}
	\end{minipage}}
  {
	\begin{minipage}[c][\width]{
	   0.5\textwidth}
	   \centering
	   \includegraphics[width=\textwidth]{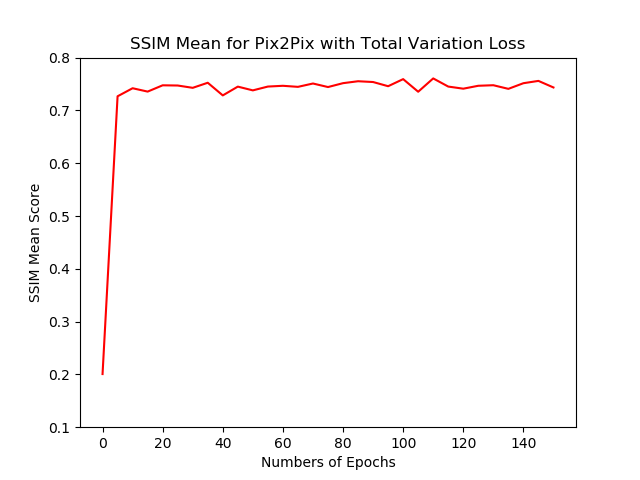}
	\end{minipage}}
\caption{SSIM average scores of C-GAN models trained to epochs 0-150 \\ Pix2Pix (left), modified Pix2Pix (right).}
\label{fig:ssim_mean_pix2pix_both}
\end{figure}

We then examine the learning process of the C-GAN model to analyze why C-GAN performs best out of the three models. We find that C-GAN is able to quickly overcome texture and color problems to focus on the harder part of learning detail. In Figure \ref{fig:ssim_mean_pix2pix_both}, we see that the SSIM index improves quickly until around epoch 10 and then gradually levels out. The SSIM, which mainly focuses on the texture of the image, is only good at distinguishing bad texture (i.e. grainy or grid-shaped coloring). Thus, C-GAN focuses on learning to output the right texture from the start to epoch 10. Then, in Figure \ref{fig:fid_pix2pix_both}, the FID improves quickly until approximately epoch 35 and then gradually levels out. The FID is able to detect both "texture" and "color" problems, so from epoch 10 to epoch 35, the C-GAN mainly focuses on improving its skills at coloring one area with one single color (as opposed to two colors smudged together).  Lastly, the FID and SSIM scores improve slightly from epoch 35 onwards, even though qualitatively, the images show improvement until approximately epoch 100. This is because the GAN focuses on learning small details after epoch 35 that the two metrics do not weight heavily. As mistakes in the small details are not as blatant as large mistakes such the as "texture" and "color" problems, improving in those areas might seem substantial qualitatively but may be obscured by noise in the plot of the metrics.

\begin{table}
  \centering
  \begin{tabular}{llll}
    \toprule
    \cmidrule(r){1-2}
    Model     & FID     & SSIM (mean)  & SSIM (standard deviation) \\
    \midrule
    Neural Style Transfer & 345.506  & 0.6547214 & 0.09885219 \\
    CycleGAN    & 272.619 &  0.7238495  & 0.08240251  \\
    \textbf{C-GAN (Pix2Pix)}    & \textbf{227.948} & \textbf{0.7468922}  & \textbf{0.07413621}  \\
    \textbf{C-GAN (Pix2Pix modified) }    & \textbf{220.499} & \textbf{0.75587333}  & \textbf{0.07380538}  \\
    \bottomrule
  \end{tabular}
  
  \caption{Performance comparison between all models.}
  \label{fig:table_fid_ssim}
\end{table}

\section{Conclusion and Future Work}

In summary, we implement the baseline algorithm of Neural Style Transfer and the generative models of CycleGAN and C-GAN. Our models produce decent outcomes on this anime colorization task, with C-GAN yielding the best performance as it improves past its texture and color problems. Additionally, the use of the total variation loss is of great help in removing noise and high frequency artifacts in the final outputs. Such a deep learning program has many practical applications, specifically in designing colorized images at an accelerated pace and bringing to life stunning manga and anime products.

The next step would be to fine-tune hyperparameters to improve our current models and to design a real-fake experiment (similar to the one in \cite{cGAN2}) in order to test their performance based on human perception. Additionally, we want to utilize the higher resolution of $512 \times 512$ to generate high-quality outputs that can be deployed in the real industry. We would also experiment with various generative models such as other GANs and conditional variational autoencoders as well as modify the network architectures, such as employing ResNet, ImageGAN and so on. Other options include doing transfer learning with pre-trained weights for Pix2Pix and harnessing different color spaces. Finally, one feature we are excited to implement is to condition on certain colors for the image in order to give the user more control. \cite{taivu1} \cite{taivu2}

\newpage

\begin{figure}[ht]
  {
	\begin{minipage}[c][\width]{
	   0.5\textwidth}
	   \centering
	   \includegraphics[width=\textwidth]{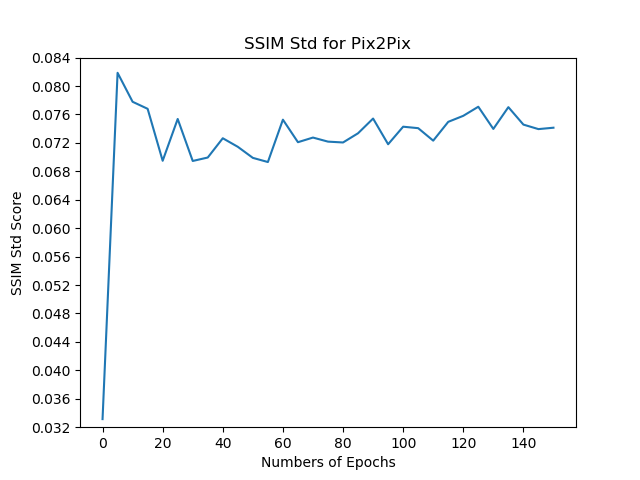}
	\end{minipage}}
  {
	\begin{minipage}[c][\width]{
	   0.5\textwidth}
	   \centering
	   \includegraphics[width=\textwidth]{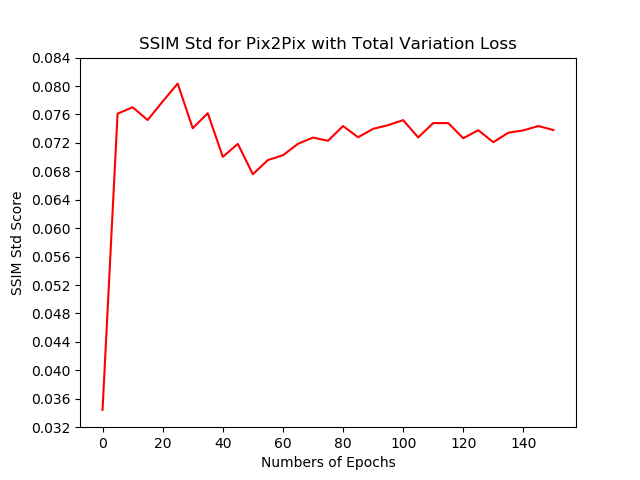}
	\end{minipage}}
\caption{SSIM standard deviations of C-GAN models trained to epochs 0-150 \\ Pix2Pix (left), modified Pix2Pix (right).}
\label{fig:ssim_std_pix2pix_both}
\end{figure}
\nocite{*}
\newpage

\bibliographystyle{ieeetr}
\bibliography{main}

\end{document}